%% file: main.tex
\documentclass{ecai}
\pdfoutput=1
\usepackage{times}
\usepackage{graphicx}
\usepackage{latexsym}
\usepackage{amsfonts}       

\ecaisubmission   
\usepackage[utf8]{inputenc} 
\usepackage[T1]{fontenc}    
\usepackage{hyperref}       
\usepackage{url}            
\usepackage{booktabs}       
\usepackage{amsfonts}       
\usepackage{nicefrac}       
\usepackage{microtype}      
\usepackage{multirow}
\usepackage{graphicx}
\usepackage{textcomp}
\usepackage{subfig}
\usepackage{xcolor}
\usepackage{xspace}
\usepackage[vlined,ruled,linesnumbered]{algorithm2e}

\newcommand{\algns}{QActor }
\newcommand{\alg}{\algns\xspace}

\begin{document}

\title{QActor: On-line Active Learning for Noisy Labeled Stream Data}

\author{Paper \#437}

\author{Taraneh Younesian \institute{TU Delft,
 The Netherlands, email: t.younesian@tudelft.nl} \and Zilong Zhao \institute{Universit\'e Grenoble Alpes, France, email: zilong.zhao@gipsa-lab.fr}  \and
 Amirmasoud Ghiassi \institute{TU Delft,
 The Netherlands, email: s.ghiassi@tudelft.nl} \and
 Robert Birke \institute{ABB Future Labs, Switzerland. email: robert.birke@ch.abb.com} \and Lydia Y. Chen\institute{TU Delft,
 The Netherlands, email: lydiaychen@ieee.org} }

\maketitle
\bibliographystyle{ecai}

\begin{abstract}
  Noisy labeled data is more a norm than a rarity for self-generated content that is continuously published on the web and social media. Due to privacy concerns and governmental regulations, such a data stream can only be stored and used for learning purpose in a limited duration.
  To overcome the noise in this on-line scenario  we propose \alg which novel combines: the selection of supposedly clean samples via quality models and actively querying an oracle for the most informative true labels.  
While the former can suffer from low data volumes of on-line scenarios, the latter is constrained by the availability and costs of human experts.
\alg swiftly combines the merits of quality models for data filtering and oracle queries for cleaning the most informative data. The objective of \alg is to leverage the stringent oracle budget to robustly maximize the learning accuracy. 
  \alg explores various strategies combining different query allocations and uncertainty measures. A central feature of \alg is to dynamically adjust the query limit according to the learning loss for each data batch. We extensively evaluate different image datasets fed into the classifier that can be standard machine learning (ML) models or deep neural networks (DNN) with noise label ratios ranging between 30\% and 80\%.
  Our results show that \alg can nearly match the optimal accuracy achieved using only clean data at the cost of at most an additional 6\% of ground truth data from the oracle.
\end{abstract}

\input{ECAI-Introduction}
\input{IJCAI-Related}

\input{ECAI-Method}
\input{ECAI-Evaluation}


\section{Conclusion}
In this paper, we consider a challenging problem that the data arrives in stream with limited validity and its label quality varies. We propose \alg, an on-line learning algorithm for very noisy label datasets. The core of \alg is composed of a quality model that filters out the noisy labels and an active learner that smartly selects noisy instances to be relabeled by an oracle. The unique feature of \alg is to explore different uncertain measures and dynamically allocate queries in batches based on the learning loss in an on-line fashion. The flexible design enables \alg to be generalized on both standard and deep learning models that have limited clean data labels.  Our extensive evaluation on five datasets show that \alg can effectively combine the merits of the quality model and active learning when encountering streaming extremely noisy labels, i.e., up to 80\%.   \alg can achieve similar classification accuracy as the cases without any noisy labels, i.e., with roughly 6\% and 10\% of oracle information for standard ML and DNN, respectively. For future work, we plan to explore various measures of uncertainty and different types of DNN models which can either provide more accurate predictions on probability, e.g., Bayesian neural networks.

\begin{table}
\caption{Comparison against model baselines with $30\%$ noise in the online setting.}
\label{tab:baselines_table}
\centering
\resizebox{0.85\columnwidth}{!}{%
\begin{tabular}{c|l|c|c}
\cline{2-4}
 \multicolumn{1}{c}{} & \multirow{2}{*}{\textbf{Methods}} & \multicolumn{2}{c}{\textbf{Accuracy (\%)}}\\
\cline{3-4}
\multicolumn{1}{c}{} & & CIFAR-10 & CIFAR-100\\
\hline \hline
\multirow{5}{*}{\rotatebox[origin=c]{90}{\textbf{Baselines}}}
 & D2L             & 52.9 & 11.97\\ 
 & Forward         & 50.74 & 19.21\\ 
 & Co-teaching     & 60.70 & 26.20\\
 & Bootstrap soft  & 52.52 & 17.61\\
 & Bootstrap hard  & 51.79 & 23.89\\
\hline
\multirow{3}{*}{\rotatebox[origin=c]{90}{\textbf{Our}}}
 & \algns(5\%)     & 75.98 & 39.10\\
 & \algns(10\%)    & 77.73 & 36.55\\
 & \alg$^D$        & 75.78 & 38.21\\
\end{tabular}
 }
\end{table}

\bibliography{ecai}
\end{document}

%% file: ECAI-Introduction.tex
\section{Introduction}

We are in the era of big data, which are continuously generated on different web platforms, e.g., social media, and disseminated via search engines often in a casual and unstructured way. Consequently, such a big data experience suffers from diversified quality issues, e.g., images tagged with incorrect labels, so called noisy labels. Today's easy access to this large amount of data further exasperates the presence of extremely noisy data. According to~\cite{baddata}, noisy data costs the US industry more than \$3 trillion per year to cleanse or to mitigate the impact of derived incorrect analyses. While the learning models conveniently leverage such a free source of data, its quality greatly undermines the learning efficiencies and their associate utilities~\cite{jiang2017mentornet}. For example~\cite{Zhang2017memorization}, using the image classifier trained from data with highly noisy labels can significantly degrade the classification accuracy and hinder its applicability on different application domains.

Another challenge brought by big data is the stream of data generation and continuous data curation. On the one hand, this invalidates the assumptions of off-line learning scenarios and drastically increases the storage overhead. On the other hand, due to the privacy concern and government regulation, e.g., European GDPR, data shall be closely managed, imposing a limit on using curated data from the public domains. As such, today's machine learning models, e.g., classifying images, in reality often encounter such data that arrives in a stream of high velocity and can only be kept for a limited time. It is no mean feat to derive learning models which can cater to such a multi-faced challenge, i.e., noisy stream data.

Noisy label issue has been a long standing challenge~\cite{patrini2017making}, from standard machine learning (ML) models to the recent deep neural networks (DNN), whose large learning capacities can have detrimental memorization effects on dirty labels~\cite{Zhang2017memorization}. The central theme here is to filter out the suspicious data which might have corrupted labels via quality estimates. Although such approaches show promising results in combating noisy labels, the applicability to noisy stream data is unfortunately limited, due to their assumption of off-line and complete data availability. The other drawback of filtering approaches is the risk of dropping informative data points which can be influential for the underlying learning models. For instance, images with corrupted labels can be exactly on the class boundaries. It might be worthwhile to actively cleanse such data due to its high potential in improving the learning tasks, even at a certain expense.

Active learning techniques~\cite{settles2009active} are designed to query extra information from the oracle for the data whose (true) labels are not readily available. Such an oracle is assumed to know (uncorrected) labels. Due to the high oracle query cost, only the informative/uncertain data is queried within a certain query budget. The majority of active learning approaches focus on the off-line scenario and constant budget, except~\cite{vzliobaite2013active} that explores the dynamic budget based on the classification errors on one by one drifting streaming data. Motivated by its power of cleansing data, ~\cite{bouguelia2015stream} applies active learning techniques on support vector machines which encounter moderate  noise ratios, i.e., roughly 30\%. The efficacy of active learning relies on the identifying the most informative instances based on uncertainty measurements of learning tasks, e.g., class probability~\cite{schohn2000less} or entropy value~\cite{holub2008entropy}. While the related work shows promising results of active learning on noisy labels, it is not clear how active query approach can be adopted when encountering noisy data streams that can be learnt only for a short period of time.

In this paper, we focus on a challenging multi-class learning problem whose data is streamed and its labels are extremely noisy, i.e., more than half of the given labels of streaming data are wrong. Due to the privacy concerns and limited storage capacity, the data can only be stored for a limited amount of epochs, drastically shortening the data validity for training classifiers. In other words, only a small fraction of data is available for learning the model at any point in time, compared to the offline scenario. Our objective is to enhance the noise-resiliency of the underlying classifier by selectively learning from good data as well as noisy labels that are critical to train the classifier. In order to turn the noisy labels into a learning advantage, we resort to the oracle for recovering their label ground truth under a given query budget. Ultimately, we aim to optimize classification accuracy with a minimum number of oracle queries, in combating stream of noisy labels. 

To such an end, we design an on-line active learning algorithm, termed Quality-driven Active Learning (\alg), which combines the merits of quality models and typical active learning algorithms. Upon receiving new data instances, \alg first filters it via the quality model into ``clean'' and ``noisy'' categories. Second, \alg queries from the oracle the true labels of highly uncertain and informative noisy instances. 
Another unique feature of \alg is that the overall query budget is fixed but the number of queries per batch is 
dynamically adjusted based on the current training loss value. \alg uses more queries when the loss value increases to avoid incorrectly including noisy labels, and reduced otherwise. We extensively evaluate \alg on an extensive set of scenarios, i.e., noise ratios, multi-class classifier models, uncertain measures, and more importantly different data sets. Our results show that in the presence of very large label noise, i.e., up to 80\% corrupted labels, \alg can achieve remarkable accuracy, i.e., almost match the optimal accuracy obtained excluding all noisy labels, at the cost of just a small fraction of oracle information, i.e., up to 6\% oracle queried labels.

Our contributions are three fold. We design a novel and efficient learning framework, termed \alg,  whose core is the combination of quality model and on-line active learning. Secondly, we propose a dynamic learning strategy that can achieve similar results as the static one. Thirdly, we extensively evaluate the proposed \alg on an extensive set of scenarios and datasets, strongly arguing for the combination of human and artificial intelligence.

%% file: IJCAI-Related.tex
\section{Related Work}


\subsection{Quality model}

Human error and careless annotators results in unreliable datasets with mistakes in labels available in public domains~\cite{yan2014learning,blum2003noise}. Adversaries are another source of label noise attacking the performance of (deep) learning systems~\cite{Szegedy2014intriguing,goodfellow2014explaining}. Learning with noise in the labels with no quality filtering shows the effect of noise in degradation of the classification accuracy of deep neural networks~\cite{Zhang2017memorization}.
As mentioned in~\cite{Zhang2017memorization}, the accuracy of using trained AlexNet to classify CIFAR10 images with random label assignment drops from 77\% to 10\% due to network memorization of the noisy samples.
In Co-teaching~\cite{han2018co} two neural networks are trained simultaneously on two different data and exchange the model information trained by the data causing the lowest loss. On the other hand, the study in \cite{patrini2017making}, Forward, assumes there is a noise transition matrix to cleanse the noisy labels for deep neural network. Furthermore, D2L \cite{ma2018dimensionality} uses the Local Intrinsic Dimension (LID) as a measure to filter the noisy labeled instances during training.

\subsection{Active learning}
Active learning has been employed with a growing rate in recent studies with deep networks due to the expenses of large dataset collection. Various studies focus on the identification of informative data instances. The studies in \cite{sener2017active,sener2017geometric} consider geometrical approaches to select the data instances, i.e. the core-set, that is the representative of the data space. Meanwhile, \cite{gal2017deep} uses deep Bayesian neural networks with monte-carlo dropout to identify the most uncertain samples for labelling. Furthermore, \cite{stanitsas2017active} uses the probability output of the convolutional neural network to label the instances based on discrete entropy and best-vs-second-best.

\subsection{Online learning}

In the online setting data arrives in a conceptually infinite stream and the opportunity to learn from each sample is brief. Many papers study incremental learning algorithms to allow to learn models progressively from new data~\cite{neucom18losing,SahooPLH18IJCAI}. Only few consider noisy stream data~\cite{zhu2006effective,chu2004adaptive}.
However these studies consider ensemble of several classifiers to detect the noisy samples which is not scalable to large image datasets and deep neural networks. 




%% file: ECAI-Method.tex
\section{Quality-driven Active Learning \alg}
\subsection{Problem Statement}
We consider multi-class classification problems that map data inputs $x$ of $D$ features into labels $y$ of $C$ classes, $\mathbf{x} \in \mathbf{X}^{N \times D} $ into $y \in \bf{C}=\{1, \dots, \textit{C}\}$. A small set of initial data instances with clean labels is given, together with a testing set. A clean data instance refers to samples whose given label is properly annotated, without any alteration. A noisy data instance refers to samples whose label is corrupted, i.e. different from the true label.
Data instances are continuously collected over time. Specifically, we introduce $i$ and $t$ to denote a particularly data instance $i$ of data arriving at interval $t$ as $\mathbf{x}(t)_i$. For simplicity, we omit $t$ when referring $\mathbf{x}$. Algorithm \ref{alg:aqlpolicy} shows the overview of our proposed method.

 \begin{figure}[tp]
    \centering
    \includegraphics[width=\columnwidth]{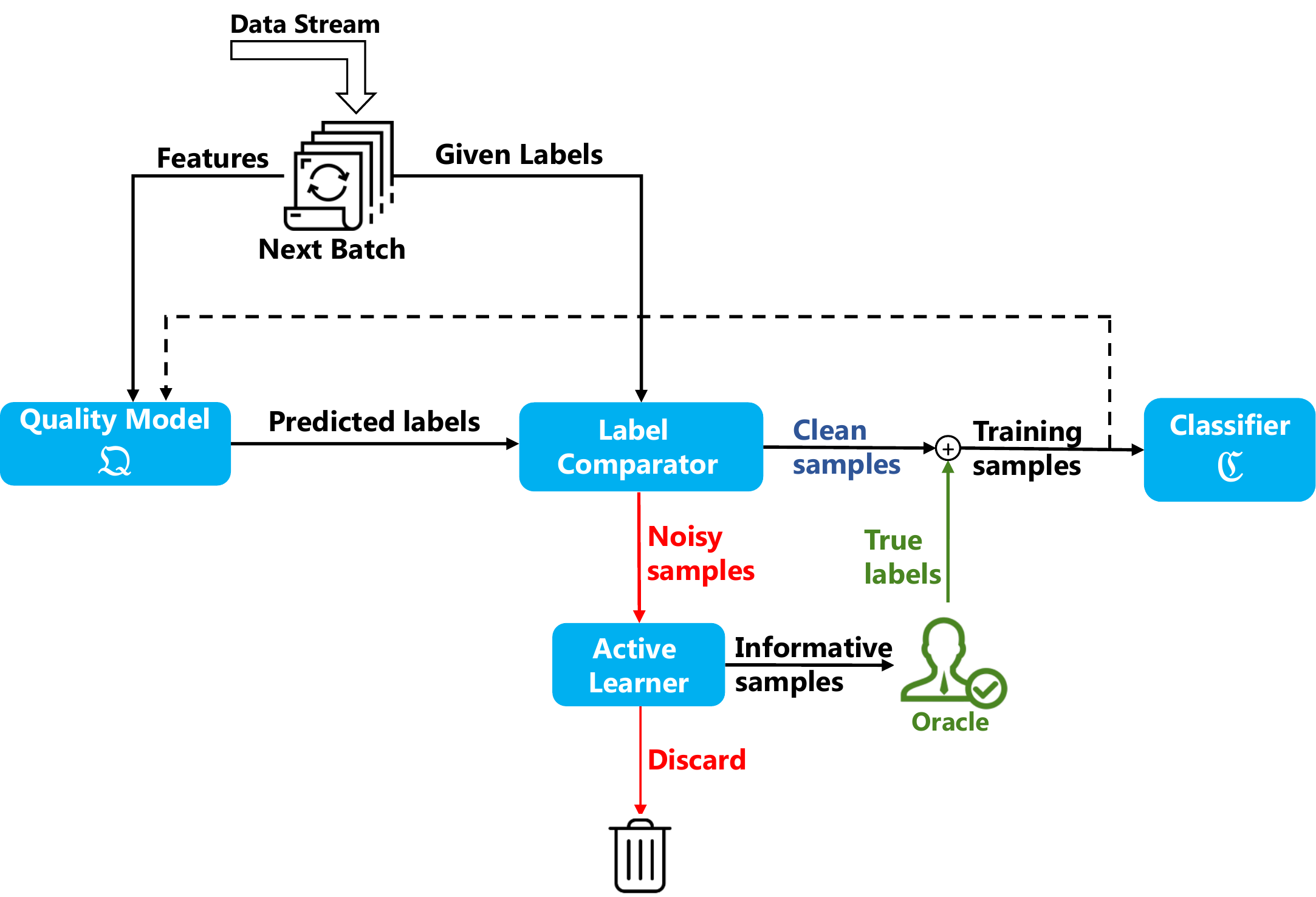}
    \caption{Overview of \alg: workflow of quality model, active learner, and classifier.}
    \label{fig:method}
\end{figure}

\subsection{Architecture of \alg}
Fig.~\ref{fig:method} depicts the architecture. The main components are the quality model $\mathfrak{Q}: \mathbf{x}\rightarrow y \in \mathbf{C}$, the label comparator which discerns noisy from clean labels, the active leaner which determines which and how many data instances to send to the oracle, and the classifier $\mathfrak{C}: \mathbf{\tilde{x}}\rightarrow y \in \mathbf{C}$. $\mathbf{\tilde{x}}$ is a subset of $\mathbf{x}$ defined below.

At each time $t$ the incoming data $x$ is first sent to the quality model from the previous time instance $\mathfrak{Q}(t-1)$  which predicts their labels. If the predicted labels are the same as the given ones, the label comparator marks them and their labels as clean, denoted as $\mathbf{x}^{cl}$; otherwise as noisy, $\mathbf{x}^{ny}$.
Noisy data instances are sent to the active learner to rank them and select $o(t)$ data instances $\mathbf{x}^{or}$ to send to oracle to query their true label. $o(t)$ is limited by the given available budget $B$, i.e., $\sum_t o(t) \le B$. Since $o(t)$ is typically much smaller than the number of noisy data instances, i.e. $o(t) \ll |\mathbf{x}^{ny}|$, instances are ranked and selected based on a uncertainty metric.
The clean and oracle relabeled instances are denoted by $\mathbf{\tilde{x}} = \mathbf{x}^{ny} \cup \mathbf{x}^{or}$ and used to re-train $\mathfrak{C}(t)$ and $\mathfrak{Q}(t)$. To avoid pitfalls in learning we monitor the accuracy on a small hold-out of the initial set. If performance drops by more than $a$ we roll back the model before processing the next data batch.

\subsection{Uncertainty Measure}
\label{ssec:uncertainty}
\subsubsection{Standard Machine Learning}
The specific quality model considered here is a multi-class SVM, 
based on a so-called decision function $f:\mathbf{x} \rightarrow N\times C$ to map data instances to classes in a one-vs-rest setting. Specifically, $f(\mathbf{x})_{i,c}$ corresponds to the decision value of a data instance $i$ for class $c$. The higher the value of $f_{i,c}, \, \forall c$ is the more certain the data instance belongs to class $c$. Hence, the uncertainty can be measured by the absolute values of $f_{i,c}$.
We consider the following uncertainty metrics for each data instance $i$ in the potentially noisy data $\mathbf{x^{ny}}$: 

\begin{enumerate}
  \item \textbf{Least Confident (LC):} where the method queries the instance that the model is the least confident to classify, i.e. $\min_c {|f(\mathbf{x_i^{ny}})|}$.
  \item \textbf{Best-versus-second-best (BvSB)~\cite{joshi2009multi}:} i.e., $|f_{best}(\mathbf{x_i^{ny}})|-|f_{second-best}(\mathbf{x_i^{ny}})|$. This method considers the difference between the decision functions of the two classes having the highest values as the uncertainty measure.
\end{enumerate}


\subsubsection{Deep Neural Networks} 
We use Deep Neural Networks (DNN) as classifier for this part since they have shown extremely promising results in classifying complex image datasets~\cite{goodfellow2014explaining}. Due to the high training costs of deep neural networks, to reduce the computational burden, instead of having two different models for $\mathfrak{Q}$ and $\mathfrak{C}$, we leverage the time difference between data arrivals to use the previously trained $\mathfrak{C}$ as $\mathfrak{Q}$ for the following time period, i.e., $\mathfrak{Q}(t) = \mathfrak{C}(t-1)$. This optimization allows us to train only one model per time period.

We estimate the prediction uncertainty of DNN via the prediction probabilities $P$ at the output of the softmax layer. In general the higher the probability the more confident the prediction, but also how spread out the probability distribution is counts. Hence, we test three different uncertainty metrics: two drawn from the classic active-learning literature, plus one based on the training loss function of the DNN. Here, we use $P_{best}$ and $P_{second-best}$ to denote the probabilities of the most likely and second most likely class.

\begin{enumerate}
  \item \textbf{Least Confident (LC)} queries the instance where the model has the lowest classification confidence, i.e. $\min {P_{best}(\mathbf{x_i^{ny}})}$.
  \item \textbf{Best-versus-second-best (BvSB)}~\cite{joshi2009multi} considers instances where the difference between the two most likely classes is minimum meaning that these instances could have easily been classified either way, i.e. $\min P_{best}(\mathbf{x_i^{ny}})-P_{second-best}(\mathbf{x_i^{ny}})$.
  \item \textbf{Highest Loss (HL)} considers the instances which produce the highest values in the loss function $\mathcal{L}$ when compared to their given label meaning they are the hardest to learn, i.e. $\max \mathcal{L}(x_i^{ny}, y_i)$.
\end{enumerate}

\subsection{Active Learner Query Policies}
The aforementioned uncertainty measures are used by the active learner in combination with two different  policies on how to use the query budget over time:

\begin{enumerate}
    \item[] \textbf{Static policy}. The active learner asks a constant number $o(t) = M, \forall t$ of queries at every batch arrival of stream data. Essentially, for each batch, the active learner queries the most uncertain $M$ data instances that are considered noisy by the quality model.

    \item[] \textbf{Dynamic policy}. The active learner dynamically adjusts $o(t)$ based on the loss function of the quality model. The rationale behind is to increase the number of queries when the quality model has a low learning capacity, reflected by high loss function values, and to decrease the number of queries when loss function converges to lower values.  Specifically, we propose to adjust $o(t)$ as following:
\begin{equation} \label{eq:1}
    o(t)=o(t-1)(1-\frac{L^{\mathfrak{Q}_{t-2}}-L^{\mathfrak{Q}_{t-1}}}{L^{\mathfrak{Q}_{t-1}}})
\end{equation}
where
\begin{equation} \label{eq:2}
    L^{\mathfrak{Q}_{t-1}}=-\frac{1}{N}\sum_{i=1}^N\sum_{c=1}^C p(y=c|\mathbf{x}_i) \log p(y=c|\mathbf{x}_i)
\end{equation}
is the loss function based on entropy. We further note the number of active queries is subject to the budget constraint $B$, i.e. the number of active queries used is: 
\begin{equation}
  \min (B - \sum_{j}^{t-1}o(j),o(t))  
\end{equation}
\end{enumerate}

\begin{algorithm}[tb]

\SetAlgoLined
\SetKwInOut{Input}{Input}\SetKwInOut{Output}{Output}
\Input{Initial dataset $D^I$, Data batches $D$ made of: samples $\mathbf{x}$, given labels $\mathbf{y}$, Budget $B$}

\Output{Quality model $\mathfrak{Q}$, Classifier $\mathfrak{C}$}
\vspace{0.2cm}
Train $\mathfrak{Q}$ and $\mathfrak{C}$ with $D^I$ \\
\vspace{0.2cm}
\ForEach{arriving $D$}{
$y^p$ := Predict label by Quality model $\mathfrak{Q}$ \\ 
$\mathbf{x}^{cl}$ = $\{\forall x_i \in D,  y_i^p = y_i\}$ \\
$\mathbf{x}^{ny}$ = $\{\forall x_i \in D,  y_i^p \neq y_i\}$ \\
\vspace{0.2cm}
$x^{or}$ = Select $o(t)$ samples from $\mathbf{x}^{ny}$ according to Section 3.3 and 3.4 \\
Relabel $x^{os}$ via oracle \\
\vspace{0.2cm}
Train $\mathfrak{Q}$ and $\mathfrak{C}$ with $\mathbf{x}^{cl} \cup \mathbf{x}^{or}$
}

\caption{Quality Driven Active Learning.}
\label{alg:aqlpolicy}
\end{algorithm}

Standard active learning studies query one instance at a time and retain the model by adding that instance to the training set. Then the learner queries the next instance based on the retrained model and repeats the procedures until all the budget is spent or the desired performance is achieved. However, in an online setting where data arrives constantly in batches, often it is too computationally expensive to retrain the model after each active query and repeat for the next one. Therefore, we decided to query $o(t)$ instances at once and retrain our model just once at each batch arrival. This applies to both policies: static and dynamic. This matches well our online setting where the number of batches is big and models are retrained with the new data at each batch arrival.

%% file: ECAI-Evaluation.tex

\section{Evaluation}
\subsection{Experimental Setup}
\subsubsection{Datasets}
We consider two types of datasets. The first type of datasets represents the more standard ML approach characterised by handcrafted features. The second type instead directly uses the pixels values and represents the deep learning approach which integrates feature selection into the training process.
For the first type we use four multi-class datasets with different size and features from the \emph{UCI machine learning repository} \cite{Dua:2019}: \emph{letter}, \emph{pendigits}, \emph{usps} and \emph{optdigits}. The \emph{letter} dataset tries to identify the 26 capital letters of the English alphabet with 20 different fonts. The reaminign three target the recognition of handwritten digits via different handcrafted features and from different number of people.
For the second type we use the well-known CIFAR-10 and CIFAR-100 datasets~\cite{kriz-cifar10}. In particular we use the well-known CIFAR-10 and CIFAR-100 datasets. These datasets try to classify colored $32\times32$-pixel images into ten and hundred classes, respectively.
CIFAR-100 is more complex due to both the higher number of classes.
Table~\ref{tab:datasets} summarises the characteristics of both groups of datasets.

\begin{table}
\centering
\caption{Summary of the main properties of evalutated datasets.}
\label{tab:datasets}
\resizebox{\columnwidth}{!}{
\begin{tabular}{l|cccc|cc}
\hline
Dataset      & letter & pendigits & usps & optdigits & CIFAR-10 & CIFAR-100\\ \hline
\# classes  $~k$  & 26  & 10      & 10   & 10  & 10 &100     \\
\# features $d$ & 16 & 16         & 256  & 64    & 32x32x3 & 32x32x3   \\
\# train    & 15000  & 7494       & 7291 & 3823 & 50000 & 50000    \\
\# test     & 5000   & 3498       & 2007 & 1797 & 10000 & 10000   \\ 
\hline
\end{tabular}
}
\end{table}

\subsubsection{Label noise}
We inject label noise into the training set by corrupting the label of randomly sampled data instances. We term the sampling probability as noise rate. Corrupted samples are subject to symmetric noise, e.g., the true label is exchanged with a random different label with uniform probability. Test data is not subject to label noise.
\input{IJCAI-static_table.tex}

\begin{figure*}[th]
	\centering
	{
	\subfloat[pendigits ]{
	    \label{subfig:pendigits_acc}
	    \includegraphics[width=0.5\columnwidth]{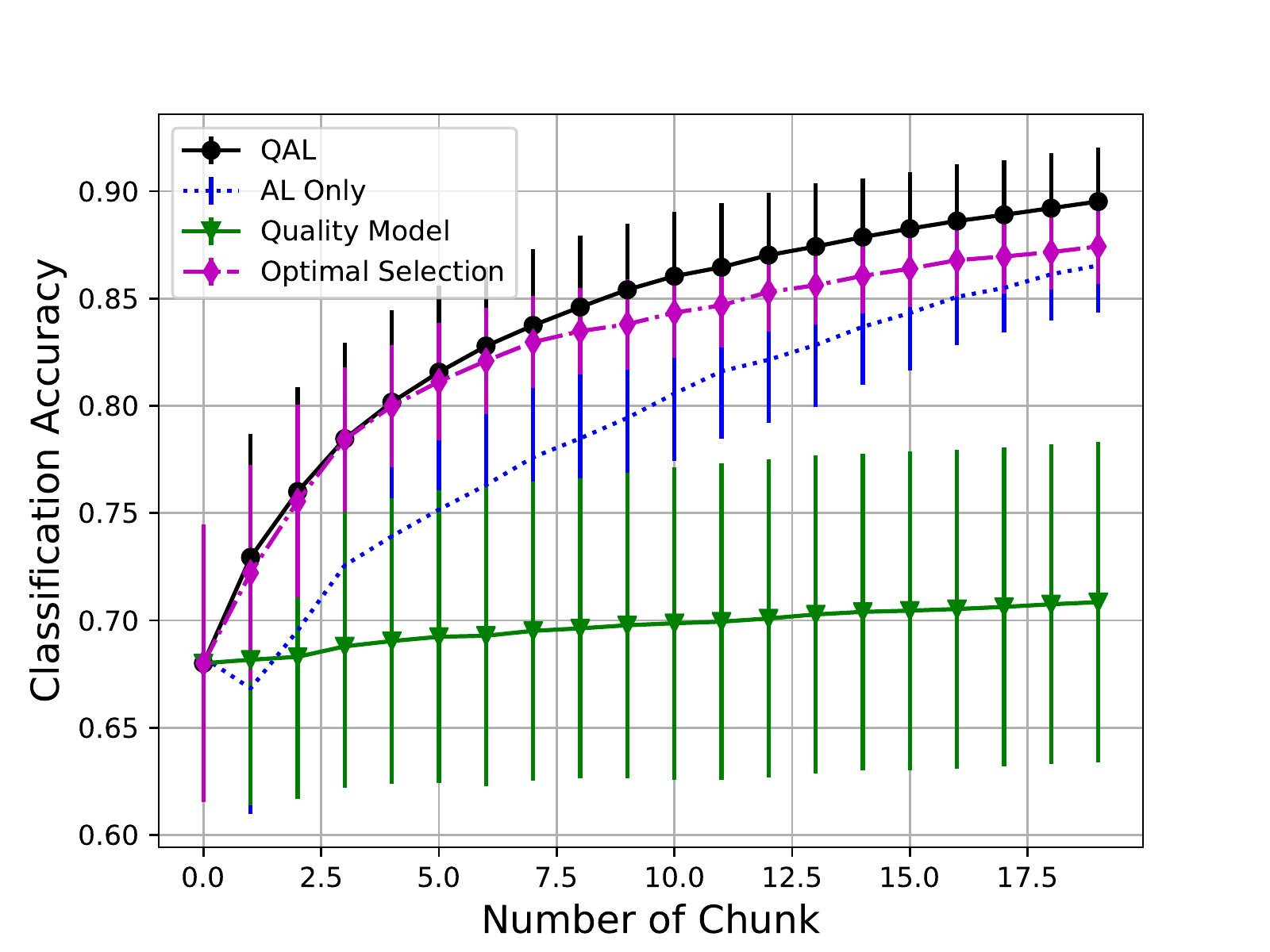}}
	\subfloat[optdigits ]{
	    \label{subfig:optdigits_acc}
	    \includegraphics[width=0.5\columnwidth]{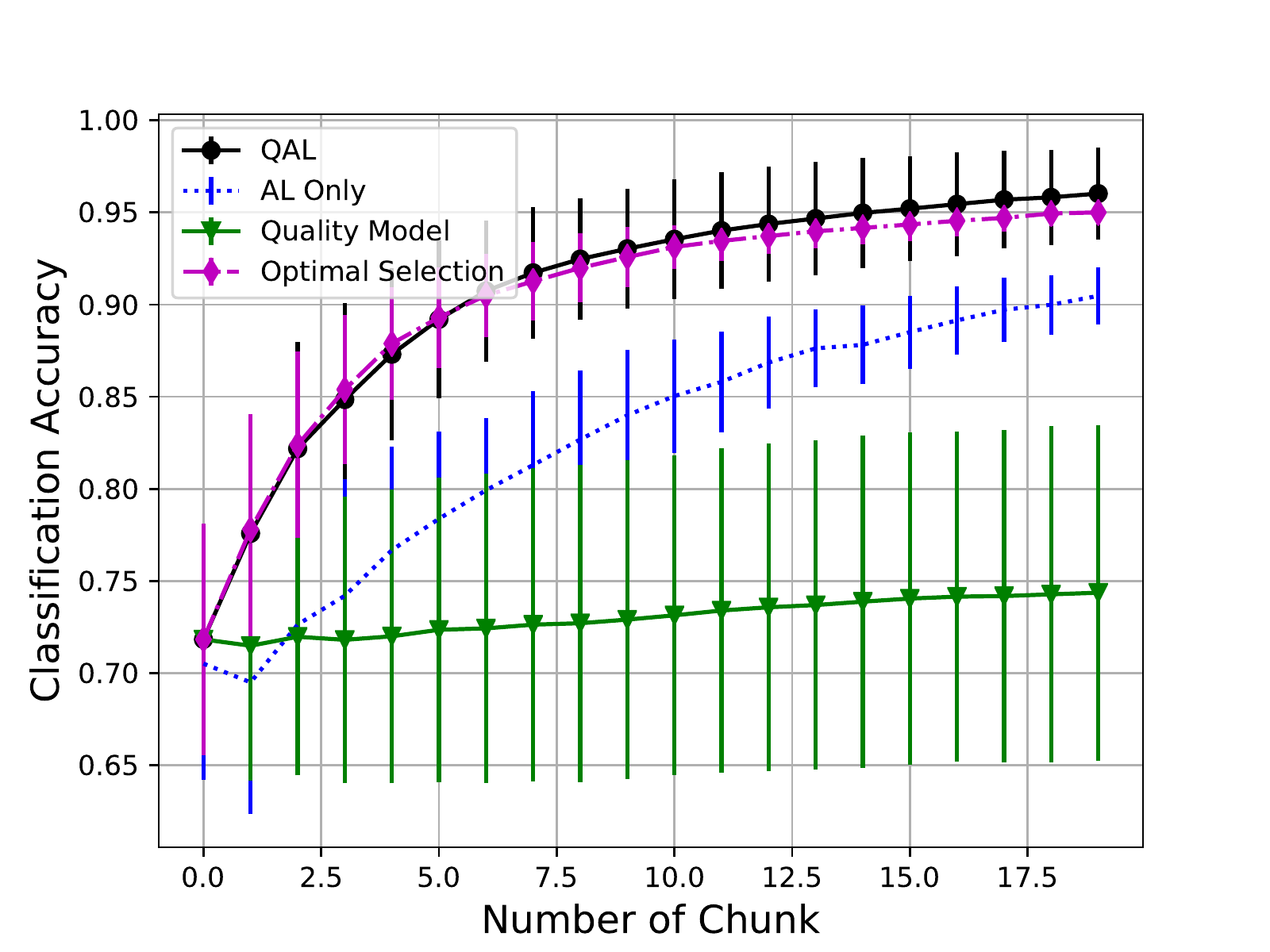}} 
	\subfloat[usps ]{
	    \label{subfig:usps_acc}
	    \includegraphics[width=0.5\columnwidth]{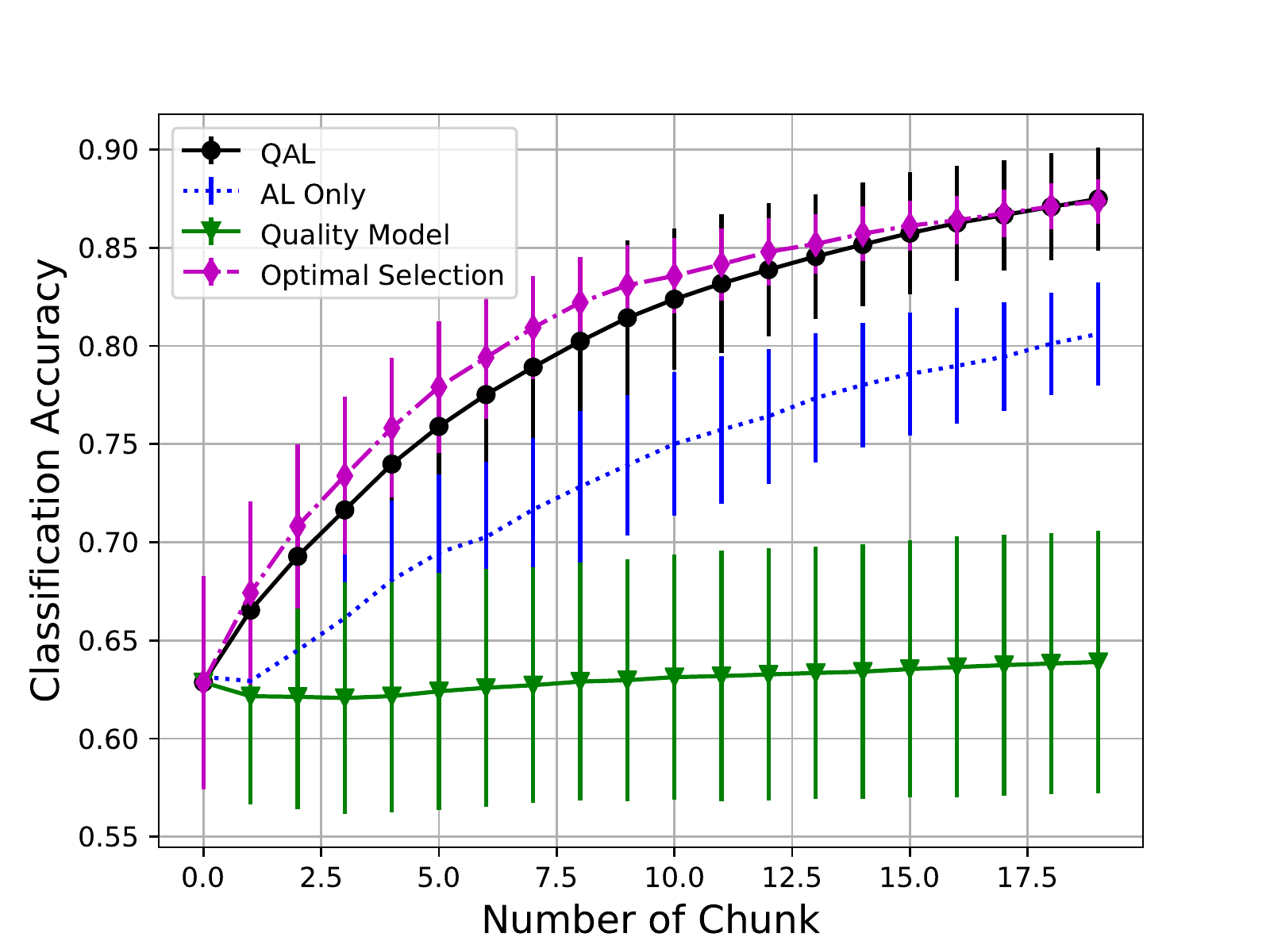}}
	\subfloat[letter ]{
	    \label{subfig:letter_acc}
	    \includegraphics[width=0.5\columnwidth]{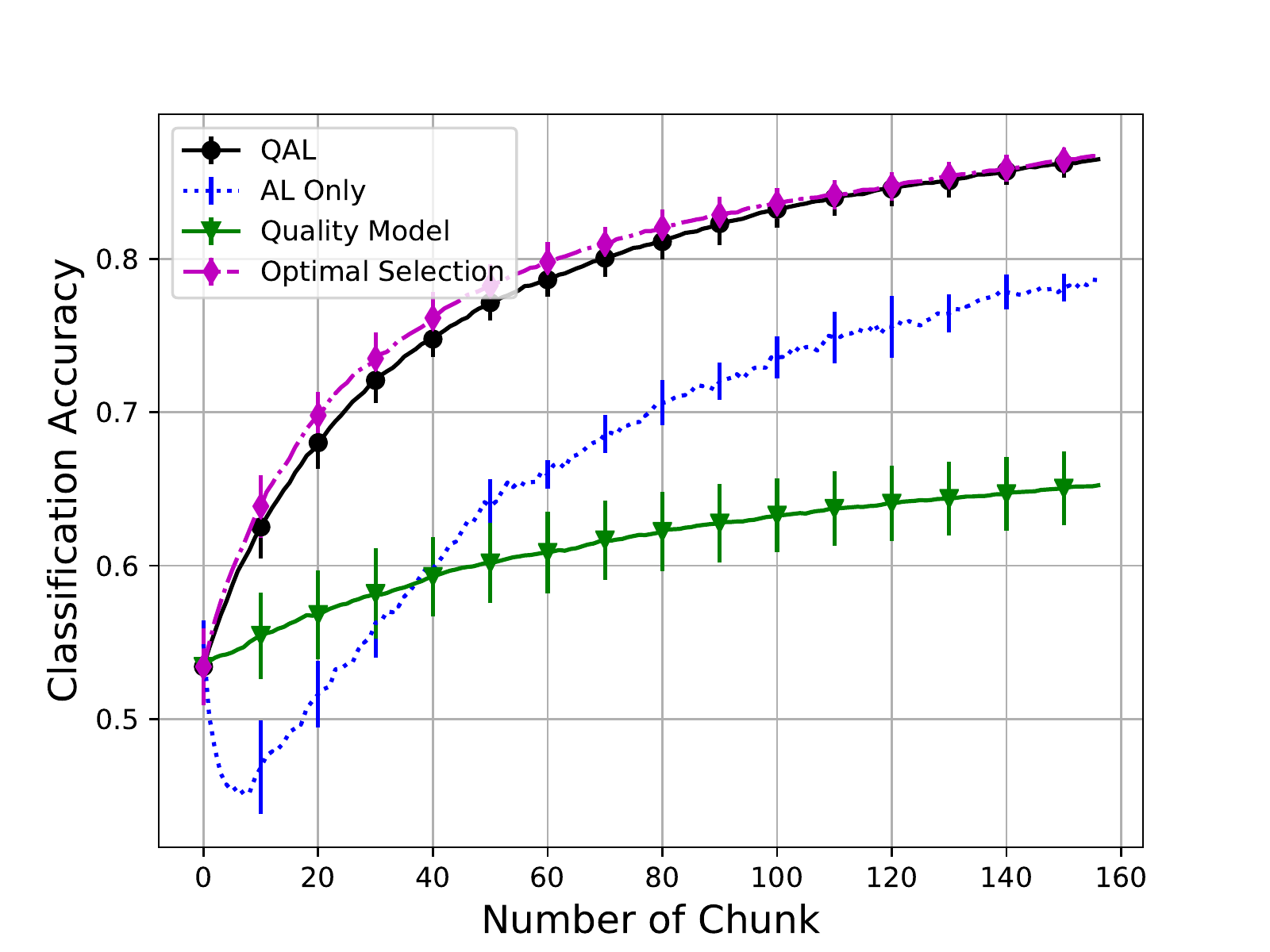}}
	}
	\caption{Classification accuracy for four data sets under $80\%$ noise ratio and 50 number of initial clean sample. Number of queries per batch is 5 for \alg and AL-only with BvSB}
	\label{fig:classification_acc_figures}
\end{figure*}

\begin{figure*}[ht]
	\centering
	\subfloat[pendigits]{
	    \label{subfig:optdigits}
	    \includegraphics[width=0.6\columnwidth]{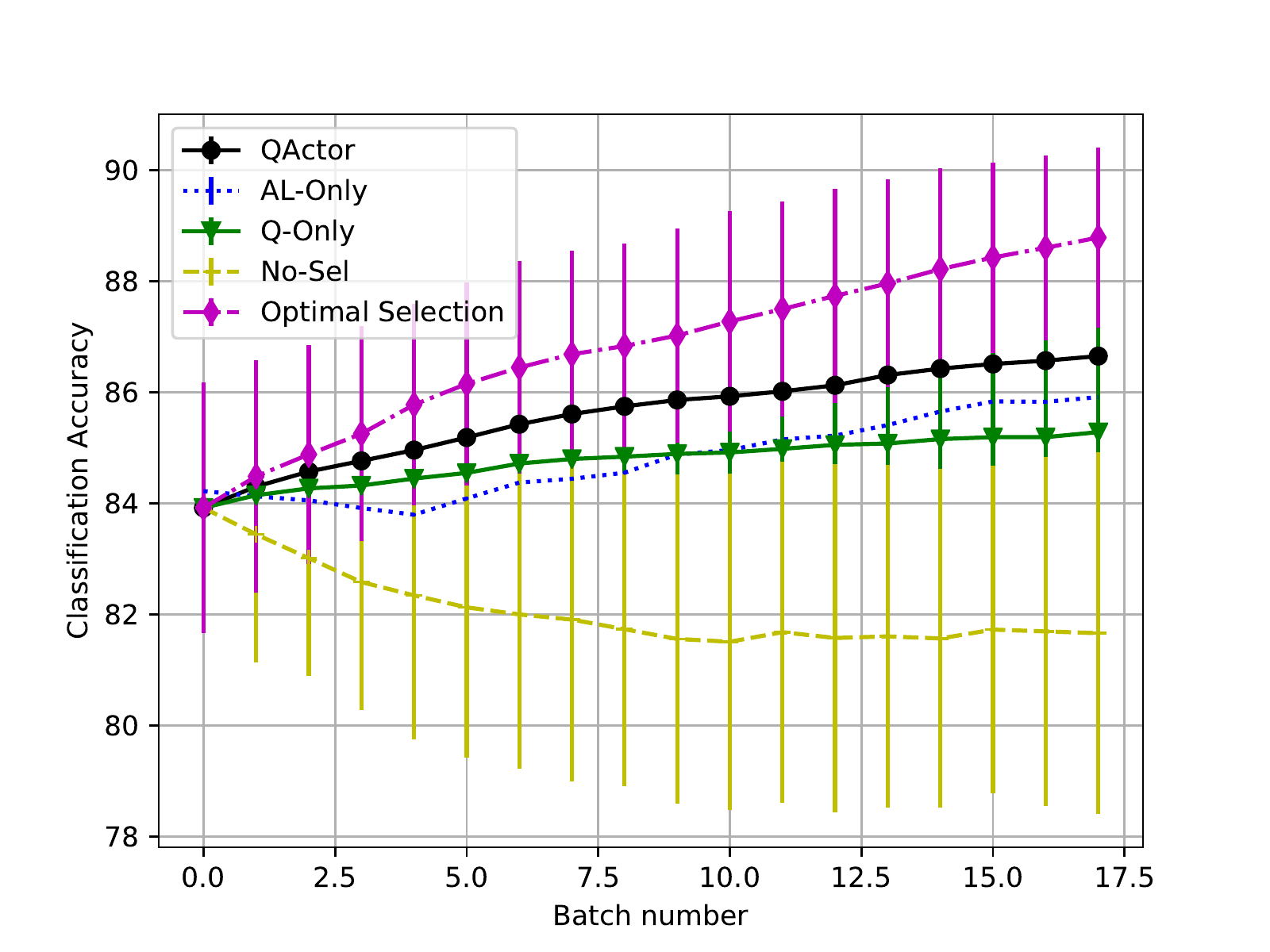}}
	\subfloat[optdigits]{
	    \label{subfig:pendigits}
	    \includegraphics[width=0.6\columnwidth]{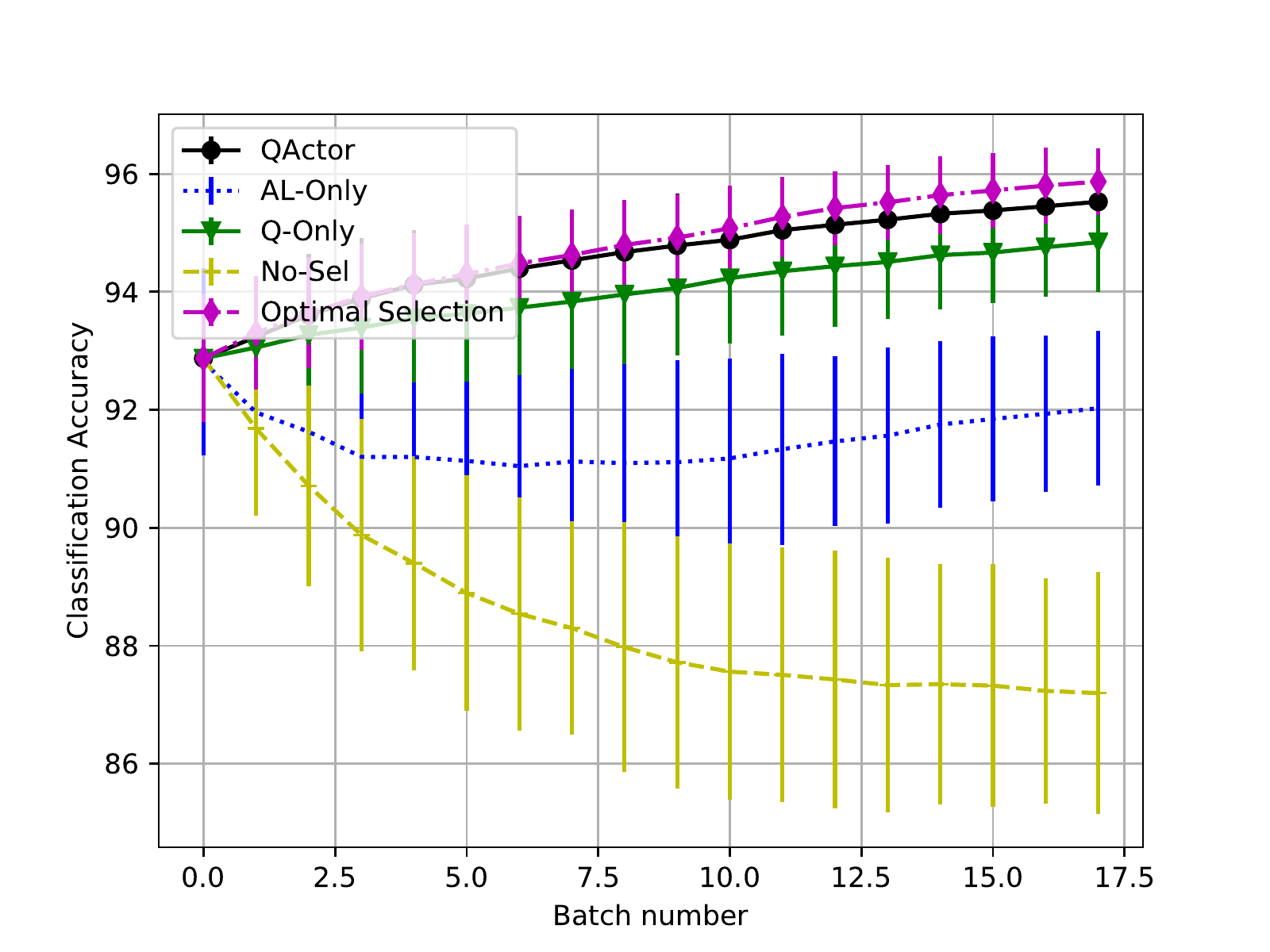}}
	 \subfloat[usps]{
	    \label{subfig:usps}
	    \includegraphics[width=0.6\columnwidth]{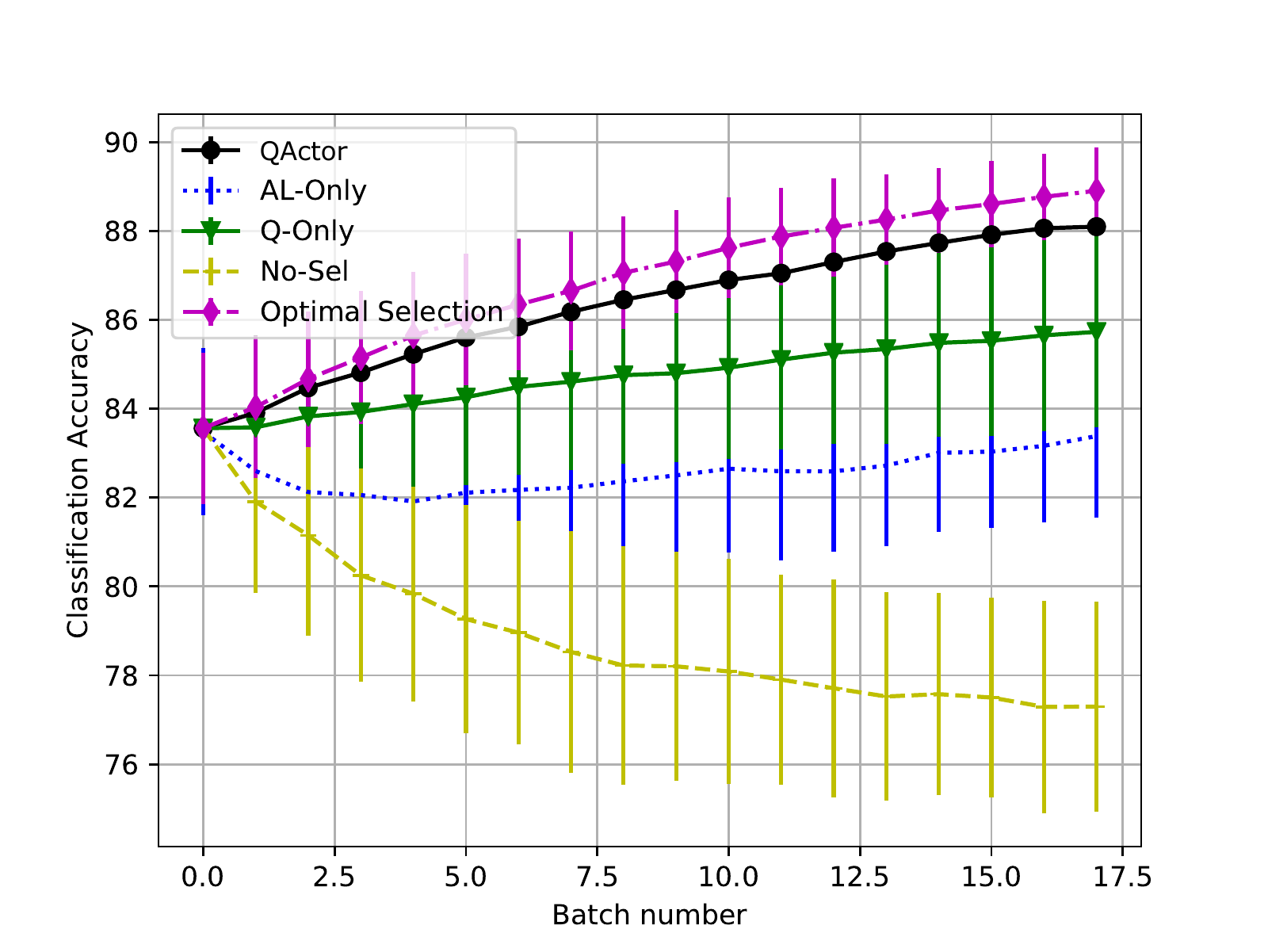}}
	\caption{Results of SVM classifier: $80\%$ noise ratio and 150 initial set with 5 queries for \alg per batch. Comparison between the proposed \alg and the baselines with uncertain measure of LC.}
	\label{fig:unc_results}
	
\end{figure*}

\subsubsection{Training Parameters for UCI datasets}
We study the effect of the number of initial clean data instances by conducting the experiments with initial clean sample set sizes of $50$ and $150$ instances for all datasets except \emph{letter}. The clean instances are chosen randomly form the trainign set.
To speed up training, we limit the data sets to $N=1050$ samples (including the initial set). For \emph{letter}, since the number of classes is higher and the dataset is more complex, we let the initial set be $150$ or $450$ clean data instances and the total training size $N=8000$. After the initial clean data batch, noisy data arrives in batches of $50$ instances. We evaluate two different constant noise rates of $60\%$ and $80\%$. 

As classifier, we try two types of standard ML techniques: random forest and SVM. 
As quality model, we use SVM since it is a well studied model to be used with active learning.
The code is written in Python using the multi-class SVM in \textit{scikit-learn}~\cite{pedregosa2011scikit}. 
Therefore the results of two methods would be different.
Under the static policy, \alg queries the true label of either the $3$ or $5$ most informative noisy samples per batch via the oracle.
Under the dynamic policy, 
\alg starts with $5$ queries per batch and adapts the number at each batch arrival as per Equation~\ref{eq:1}. 
In the results, we identify the static policy with the number of queries per batch arrival in brackets, e.g. (3), and the dynamic policy by $^D$.
We repeat each experiment $100$ times and report the average accuracy computed on the test set.

\subsubsection{Training Parameters for CIFAR-10 and CIFAR-100}
As \alg classifier for CIFAR-10 and CIFAR-100 we use the two Convolutional Neural Network (CNN) architectures defined in~\cite{wang2018iterative} with ReLU activation functions, softmax activation as image classifier and cross-entropy as loss function. We train the models by using stochastic gradient descent with momentum $0.9$, learning rate $0.01$, and weight decay $10^{-4}$. \alg and all baselines are implemented using Keras v2.2.4 and Tensorflow v1.12, except co-teaching which uses PyTorch v1.1.0.

With CIFAR-10 \alg is trained initially with clean 10000 instances and 60 epochs. The remaining 40000 instances come in batches of size 1000 with $30\%$ ($60\%$) noise. Under static policy, in each batch we query $50$ ($100$) samples, i.e. $5\%$ ($10\%$) of the batch size, actively from the oracle and retrain the model for $20$ epochs. Similarly, the dynamic policy uses $B$ equal to only $5\%$ of the total noisy data. At the end of each batch, we test the model with the test set of 10000 instances. Rollback uses $a$ = 20\%. For CIFAR-100 we increase the arriving data batch size to 10000 and 60 epochs per batch to cope with the higher complexity.
For fair comparison, baselines are also trained under the same data arrival pattern and the same CNN structure. All baselines use the same parameters as from their papers except for D2L. Here we reduce the dimensionality estimation interval to 40 and 10 for the initial and subsequent batches, respectively. This keeps roughly the original ratio against the overall training period.
For each experiment we report the average accuracy covering the last received 10000 samples.

\subsubsection{Baselines}
\label{ssec:baselines}

To better show the overall effectiveness of our proposed \alg method we compare it two sets of baselines. First we compare against different active query selection baselines:

\begin{itemize}
    \item[] \textbf{{No-Sel}}: uses all samples that arrive in the batch to train the classifier without filtering.

    \item[] \textbf{{Q-only}}: in this case the quality model filters the suspicious samples but there is no active learner to relabel the informative noisy instances. Therefore the classifier will train only on the clean data instances identified by the quality model.

    \item[] \textbf{{AL-only}}: here there is no quality model to separate the clean from suspicious data instances. The classifier is trained without the quality model filtering the data, however, informative instances identified by the active learner are relabeled by the oracle.

    \item[] \textbf{{Opt-Sel}}: which assumes a perfect quality model able to identify all the true clean and noisy samples and uses all the clean samples for training the classifier without active learning.
\end{itemize}

Second we put \alg in the context of other noise-resistant techniques drawn from the related work on learning with noise and adapted to the online scenario:

\begin{itemize}
    \item[] \textbf{D2L}~\cite{wang2018iterative}: estimates the dimensionality of subspaces during training to adapt the loss function.
    \item[] \textbf{Forward}~\cite{patrini2017making}: corrects the loss function based on the noise transition matrix.
    \item[] \textbf{Bootstrap}~\cite{iclrw2015reed}: using convex combination of the given and predicted labels for training.
    \item[] \textbf{Co-teaching}~\cite{han2018co}: exchanges mini-batches between two networks trained in parallel.
\end{itemize}



\section{Results}
We bring our results in two section for standard ML models with \emph{UCI} datasets and deep neural networks with CIFAR-10 and CIFAR-100. For each we cover the static and dynamic policies in separate sections.

\subsection{Standard Machine Learning}
\subsubsection{Static \alg}

\input{ECAI-eval_static.tex}

\input{ECAI-rf_table.tex}

\subsubsection{Dynamic \alg}

\input{ECAI-eval_dynamic.tex}

\subsection{Deep Neural Networks}
\input{IJCAI-Results.tex}

%% file: IJCAI-static_table.tex
\begin{table*}[htp]
\caption{Results of SVM classifier: learning accuracy across different datasets and noise ratio. five alternative approaches v.s. our proposed \algns using uncertain measure of BvSB.  }
\label{tab:bigtable2}
\centering
\resizebox{1.4\columnwidth}{!}{
\begin{tabular}{cccccccc}

\hline
\multicolumn{8}{c}{Initial size = 50}\\
\hline
\multicolumn{1}{c|}{\textbf{Dataset}}   & \multicolumn{1}{c|}{\textbf{Noise}} & \multicolumn{1}{c|}{\textbf{Opt-sel}}    & \multicolumn{1}{c|}{\textbf{No-sel}} & \multicolumn{1}{c|}{\textbf{Q-only}}   & \multicolumn{1}{c|}{\textbf{AL-only(3)}} & \multicolumn{1}{c|}{\textbf{AL-only(5)}} & \textbf{\algns(5)} \\ \hline
\multicolumn{1}{c|}{\multirow{2}{*}{usps}}       & \multicolumn{1}{c|}{80\%}       & \multicolumn{1}{c|}{87.35}          & \multicolumn{1}{c|}{64.47}  & \multicolumn{1}{c|}{63.79} & \multicolumn{1}{c|}{76.07}      & \multicolumn{1}{c|}{80.60}   & \multicolumn{1}{c}{\textbf{87.72}} \\
\multicolumn{1}{c|}{}                            & \multicolumn{1}{c|}{60\%}       & \multicolumn{1}{c|}{\textbf{90.20}} & \multicolumn{1}{c|}{84.89}  & \multicolumn{1}{c|}{70.32} & \multicolumn{1}{c|}{87.24}      & \multicolumn{1}{c|}{88.39}       & \multicolumn{1}{c}{89.62}     \\ \hline
\multicolumn{1}{c|}{\multirow{2}{*}{pendigits}} & \multicolumn{1}{c|}{80\%}       & \multicolumn{1}{c|}{87.60}          & \multicolumn{1}{c|}{69.88}  & \multicolumn{1}{c|}{71.82} & \multicolumn{1}{c|}{82.31}      & \multicolumn{1}{c|}{86.26}   & \multicolumn{1}{c}{\textbf{89.44}} \\
\multicolumn{1}{c|}{}                            & \multicolumn{1}{c|}{60\%}       & \multicolumn{1}{c|}{90.83}          & \multicolumn{1}{c|}{87.03}  & \multicolumn{1}{c|}{74.98} & \multicolumn{1}{c|}{89.85}      & \multicolumn{1}{c|}{91.35}      & \multicolumn{1}{c}{\textbf{91.30}} \\ \hline
\multicolumn{1}{c|}{\multirow{2}{*}{optdigits}} & \multicolumn{1}{c|}{80\%}       & \multicolumn{1}{c|}{95.09}          & \multicolumn{1}{c|}{73.20}  & \multicolumn{1}{c|}{74.72} & \multicolumn{1}{c|}{86.00}      & \multicolumn{1}{c|}{90.40}      & \multicolumn{1}{c}{\textbf{96.00}} \\
\multicolumn{1}{c|}{}                            & \multicolumn{1}{c|}{60\%}       & \multicolumn{1}{c|}{96.57} & \multicolumn{1}{c|}{93.14}  & \multicolumn{1}{c|}{81.45} & \multicolumn{1}{c|}{95.00}      & \multicolumn{1}{c|}{95.76} & \multicolumn{1}{c}{\textbf{96.93}} \\
\hline
\multicolumn{8}{c}{Initial size = 150}\\
\hline
\multicolumn{1}{c|}{\multirow{2}{*}{letter}} & \multicolumn{1}{c|}{80\%}       & \multicolumn{1}{c|}{\textbf{86.55}}          & \multicolumn{1}{c|}{66.1}  & \multicolumn{1}{c|}{65.25} & \multicolumn{1}{c|}{75.32}      & \multicolumn{1}{c|}{79.26}  & \multicolumn{1}{c}{86.35}  \\
\multicolumn{1}{c|}{}                            & \multicolumn{1}{c|}{60\%}       & \multicolumn{1}{c|}{\textbf{90.25}} & \multicolumn{1}{c|}{82.74}  & \multicolumn{1}{c|}{71.22} & \multicolumn{1}{c|}{85.44}      & \multicolumn{1}{c|}{86.62}  & \multicolumn{1}{c}{88.24} \\

\hline
\multicolumn{8}{c}{Initial size = 150}\\
\hline

\multicolumn{1}{c|}{\multirow{2}{*}{usps}}       & \multicolumn{1}{c|}{80\%}       & \multicolumn{1}{c|}{88.96}          & \multicolumn{1}{c|}{77.49}  & \multicolumn{1}{c|}{86.23} & \multicolumn{1}{c|}{83.52}      & \multicolumn{1}{c|}{85.75}      & \multicolumn{1}{c}{\textbf{90.88}}  \\
\multicolumn{1}{c|}{}                            & \multicolumn{1}{c|}{60\%}       & \multicolumn{1}{c|}{90.76} & \multicolumn{1}{c|}{86.84}  & \multicolumn{1}{c|}{88.43} & \multicolumn{1}{c|}{88.75}      & \multicolumn{1}{c|}{91.35} & \multicolumn{1}{c}{\textbf{91.39}} \\ \hline
\multicolumn{1}{c|}{\multirow{2}{*}{pendigits}} & \multicolumn{1}{c|}{80\%}       & \multicolumn{1}{c|}{89.30}          & \multicolumn{1}{c|}{81.74}  & \multicolumn{1}{c|}{86.04} & \multicolumn{1}{c|}{82.31}      & \multicolumn{1}{c|}{89.92}      & \multicolumn{1}{c}{\textbf{92.02}} \\
\multicolumn{1}{c|}{}                            & \multicolumn{1}{c|}{60\%}       & \multicolumn{1}{c|}{90.83}          & \multicolumn{1}{c|}{91.68}  & \multicolumn{1}{c|}{89.42} & \multicolumn{1}{c|}{87.48}      & \multicolumn{1}{c|}{91.16}   & \multicolumn{1}{c}{\textbf{92.90}} \\ \hline
\multicolumn{1}{c|}{\multirow{2}{*}{optdigits}} & \multicolumn{1}{c|}{80\%}       & \multicolumn{1}{c|}{95.91}          & \multicolumn{1}{c|}{87.43}  & \multicolumn{1}{c|}{94.91} & \multicolumn{1}{c|}{92.27}      & \multicolumn{1}{c|}{94.15}   & \multicolumn{1}{c}{\textbf{97.26}}  \\
\multicolumn{1}{c|}{}                            & \multicolumn{1}{c|}{60\%}       & \multicolumn{1}{c|}{96.88} & \multicolumn{1}{c|}{94.39}  & \multicolumn{1}{c|}{95.99} & \multicolumn{1}{c|}{95.92}      & \multicolumn{1}{c|}{96.46} & \multicolumn{1}{c}{\textbf{97.49}}  \\ 
\hline
\multicolumn{8}{c}{Initial size = 450}\\
\hline
\multicolumn{1}{c|}{\multirow{2}{*}{letter}} & \multicolumn{1}{c|}{80\%}       & \multicolumn{1}{c|}{87.28}          & \multicolumn{1}{c|}{70.43}  & \multicolumn{1}{c|}{79.12} & \multicolumn{1}{c|}{77.66}      & \multicolumn{1}{c|}{80.74}      & \multicolumn{1}{c}{\textbf{88.4}} \\
\multicolumn{1}{c|}{}                            & \multicolumn{1}{c|}{60\%}       & \multicolumn{1}{c|}{\textbf{90.61}} & \multicolumn{1}{c|}{84.08}  & \multicolumn{1}{c|}{82.18} & \multicolumn{1}{c|}{87.18}      & \multicolumn{1}{c|}{88.4}      & \multicolumn{1}{c}{89.57}  \\
\hline
\end{tabular}
}
\end{table*}

%% file: ECAI-eval_static.tex
Here, the number of queries per batch is constant, i.e., 3 and 5 for \algns(3) and \algns(5), respectively.
We show the results using random forest and SVM as the classifier and for BvSB as uncertainty metric in Table~\ref{tab:rf_table} and Table~\ref{tab:bigtable2}, respectively.
Preliminary results show better results for SVM than for random forest. Hence, we focus more on SVM with additional experiments for different parameter settings and all baselines.

\begin{table}
\caption{Results of \textbf{random forest} as the classifier.}
\label{tab:rf_table}
\centering
\resizebox{\columnwidth}{!}{
\begin{tabular}{c|c|c|c|c|c}
\hline
\multicolumn{6}{c}{Initial size = 50}\\
\hline
{\bf Dataset} & {\bf Noise} & {\bf Opt-Sel} & {\bf No-Sel} & {\bf Q-only} & {\bf QAL(5)} \\
\hline
\multirow{2}{*}{usps}  & 80\%  & 83.41 & 56.34 & 62.14 & {\bf83.47}\\
        & 60\% & {\bf 85.76} & 80.96 & 68.45 & 85.11 \\
\hline
\multirow{2}{*}{pendigits} & 80\% & 82.96 & 57.82 & 69.03 & {\bf 84.08} \\
    & 60\% & 84.20 & 81.74 & 72.55 & {\bf 84.44}\\
\hline

\end{tabular}
}
\end{table}

As the results in Table~\ref{tab:bigtable2} show, with small initial clean data set the quality model is not strong enough to distinguish between noisy and clean samples. In this case active learning is useful to cleanse the noisy samples. Also \alg performs better than \emph{AL-only} which shows the effect of quality model to select clean samples and filter out noisy ones. If the number of queries increases to $5$, \alg outperforms even \emph{Opt-Sel}, due to the increase in clean samples. With increase in the initial set, quality model performs very well, although our \algns(5) still outperforms all others. Figure~\ref{fig:unc_results} shows the results per batch.

We also run the experiments based on LC query selection and compared it with the baselines and different noise rates. The accuracy of the system was lower than BvSB, however our \alg was still performing better than the non-optimal baselines. Figure.~\ref{fig:unc_results} shows the accuracy over batch for three simpler datasets with initial clean set of 150 and $80\%$ noise rate. As shown, even though the initial set is big and number of active queries is high, the model performs lower than BvSB with smaller initial set. Therefore we bring all our detailed results on standard machine learning with BvSB.

%% file: ECAI-rf_table.tex


%% file: ECAI-eval_dynamic.tex
As observed earlier, the quality model is not accurate enough when starting with a small initial set.
Here we adjust the number of active queries per batch based on the loss function considering a budget. To fairly compare our proposed dynamic method with the static case, we set the budget equal to $B=60$, which is the same budget used by the static \algns(3). Our results (see \algns$^{D}$ and \algns(3)) show that this dynamic allocation leads to better performance  especially under higher noise and small initial sets, which are more challenging cases. 

\begin{figure}[th]
	\centering
	{\hfill
	\subfloat[\# active queries]{
	    \label{subfig:query_number}
	    \includegraphics[width=0.5\columnwidth]{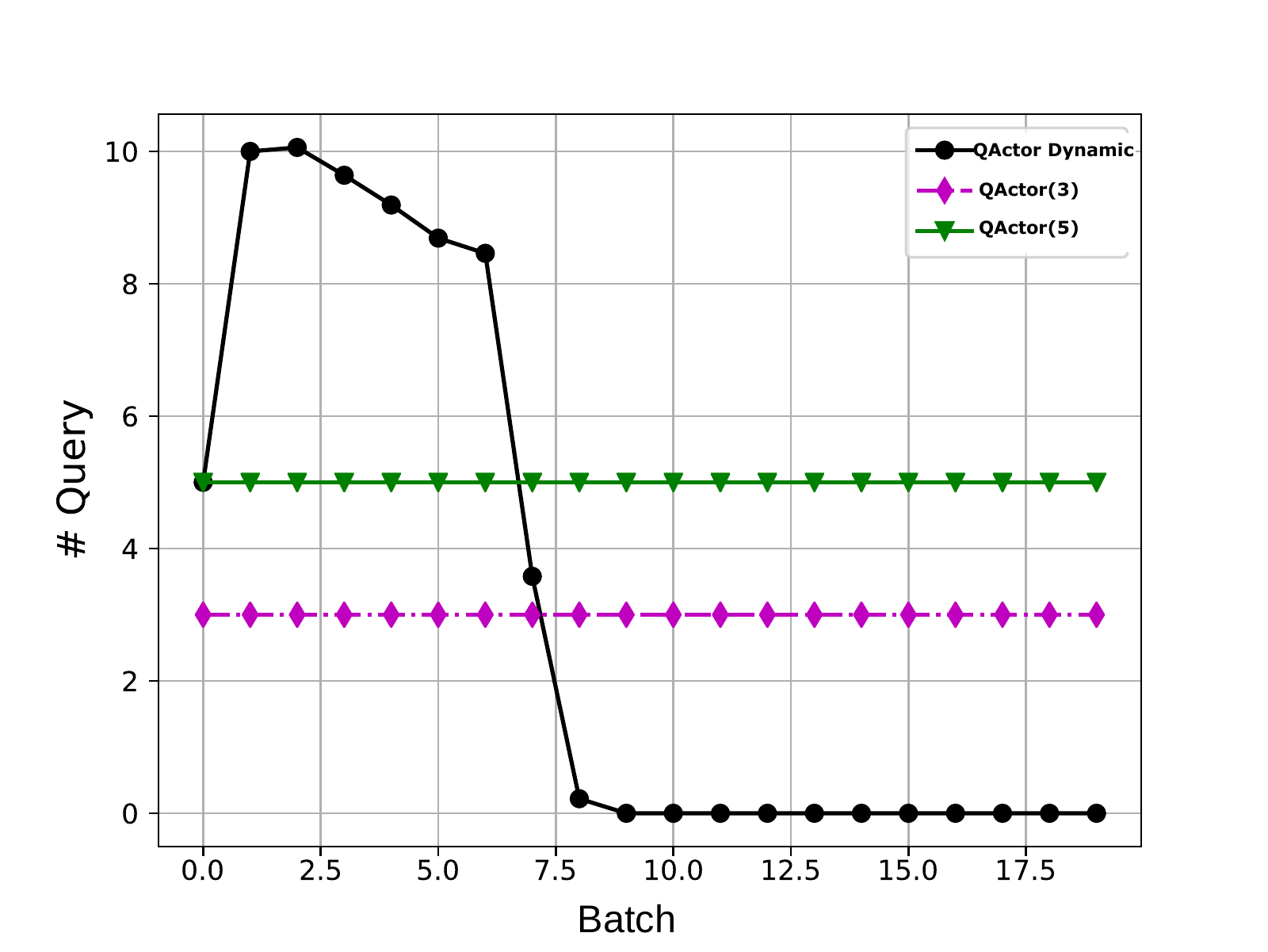}}
	\subfloat[\# training samples for the classifier, $|\mathbf{\tilde{x}}|$]{
	    \label{subfig:clean_detected}
	    \includegraphics[width=0.5\columnwidth]{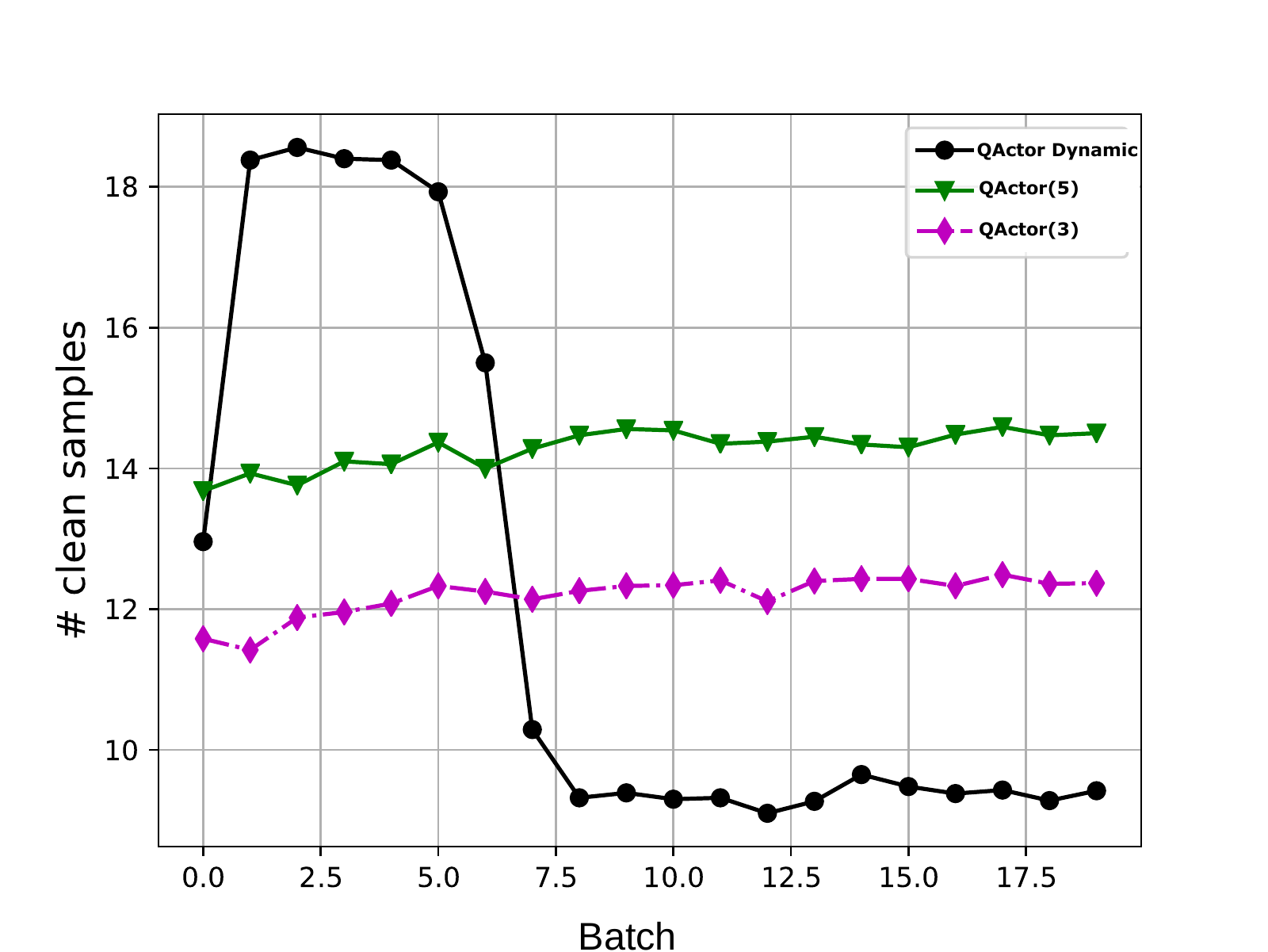}}
	\hfill}
	\caption{Results of SVM on usps: comparison of different query strategies of \alg in terms of training samples at each interval for  with initial set 50 and $80\%$ noise.}
	\label{fig:classification_acc}
\end{figure}

As is shown in Figure~\ref{subfig:query_number} which compares the number of active queries, \algns$^{D}$ increases the number of active queries at the beginning when the quality model is not yet accurate. In this case the quality model and classifier are both more vulnerable to wrong labels and we gain more from relabeled samples. Indeed Figure~\ref{subfig:clean_detected} shows that \algns$^{D}$ increased the number of clean labels at the beginning and even if afterwards this number drops, the model is now more robust and the final accuracy is better compared to the static case. Overall, \algns$^{D}$ can reach near the accuracy as static \algns(5), which has budget $B=100$.

\begin{figure*}[tp]
  \centering
  \begin{tabular}{cc}
    \hspace{-1.3cm} \includegraphics[width=1.65\columnwidth]{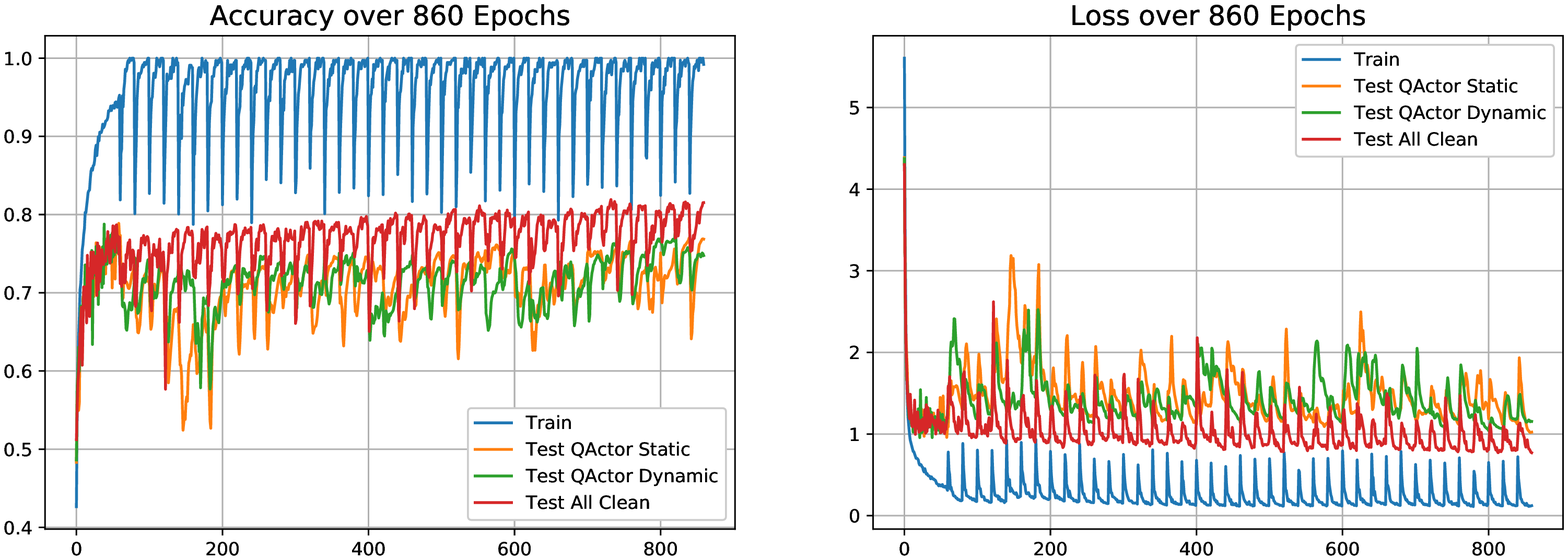}  & \hspace{-1.4cm} \includegraphics[width=0.73\columnwidth]{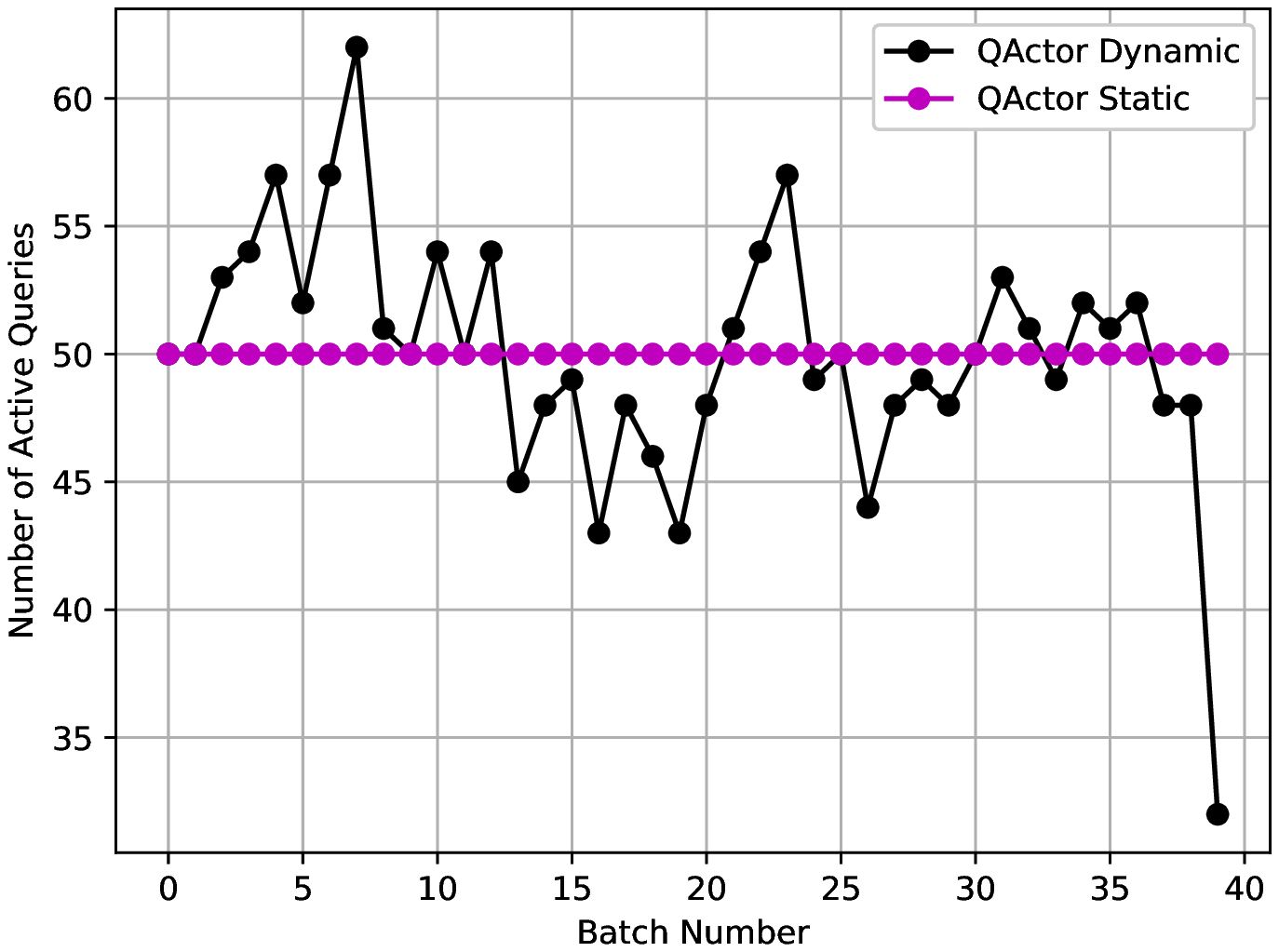}\\
      \hspace{-1.5cm} a) Accuracy \hspace{4cm} b) Loss & \hspace{-1cm} c) Number of queries
  \end{tabular}
  \caption{Comparison between \alg($5\%$) and \algns$^{D}$ on CIFAR-10 for noise rate of $60\%$ and \emph{BvSB}: a) accuracy and b) loss per epoch (20 epochs per batch of data arrival) and comparison with fully clean dataset; c) number of active queries to the oracle per batch of data arrival.}
  \label{fig:cifar-10_static_dynamic_compare}
\end{figure*}

%% file: IJCAI-Results.tex
Here we present the accuracy achieved by \alg on the CIFAR-10 and CIFAR-100 datasets. We first analyse the best uncertainty metric to use followed by our results with static and dynamic policies. Finally we conclude with a comparison against five noise-resistant model baselines from four state-of-the-art related papers.

\subsubsection{Uncertainty Metrics}



Here we compare the effect on the accuracy obtained on the test set when changing the underlying uncertainty measure used. In particular we consider the uncertainty measures laid out in Section~\ref{ssec:uncertainty}, i.e.
Least Confident (\emph{LC}), Best vs. Second Best (\emph{BvSB}) and Highest Loss (\emph{HL}), to select samples from the noisy set to be queried for their labels.
Table~\ref{tab:cifar-10-static} summarises the results on CIFAR-10 with $30\%$ and $60\%$ label noise and a constant 50 and 100 queries per data batch corresponding to $5\%$ and $10\%$ of the arriving data, respectively.

The dirtier the data, i.e. 60\% noise rate and 5\% active queries, the better performs LC (and to some extent \emph{BvSB}). \emph{LC} and \emph{BvSB} select the most uncertain samples based on uncertainty either in absolute terms, i.e. \emph{LC}, or relative terms, i.e. \emph{BvSB}. These can be samples close to either side of a decision boundary. As a consequence active queries hit both clean and noisy samples. For example, for $30\%$ noise and $5\%$ active queries, we query about $40\%$ noisy samples.
The cleaner the data, i.e. 30\% noise rate and 10\% active queries, the better performs \emph{HL}.  \emph{HL} chooses the samples which have the highest loss.
These are either intrinsically difficult samples or noisy samples where the label has been flipped and does not fit well into the space of samples of the same class. Indeed, \emph{HL} obtains an almost perfect score in actively querying only noisy samples ($>98\%$). However, as noise increases, \emph{HL} quickly becomes too conservative in selecting clean samples dropping even clean training data via the label comparator. For example, with $5\%$ queries \emph{HL} dropped on average $65.8\%$ of the arriving data even when the noise rate is only $30\%$.

In the following to compare against selection baselines and static v.s. dynamic budget allocation, we use \emph{BvSB} as uncertainty metric.

\begin{table}
\caption{Summary of learning accuracy of static policy under different uncertainty metrics and noise ratios for CIFAR-10.}
\label{tab:cifar-10-static}
\centering
\resizebox{0.8\columnwidth}{!}{
\begin{tabular}{l|c|c|c}
\hline
\multicolumn{1}{c|}{\multirow{2}{*}{\textbf{Uncertainty Metric}}} & \multicolumn{1}{l|}{\multirow{2}{*}{\textbf{Noise rate}}} & \multicolumn{2}{c}{\textbf{Query \%}} \\ \cline{3-4} 
\multicolumn{1}{l|}{} & \multicolumn{1}{l|}{} & 5\% & 10\% \\ \hline \hline
\multirow{2}{*}{Least Confident} & 30\% & 75.68 & 75.57 \\
 & 60\% & 72.13 & 75.73 \\ \hline
\multirow{2}{*}{Best vs. Second Best} & 30\% &75.98 & 74.00 \\
 & 60\% & 70.71 & 74.07  \\ \hline
\multirow{2}{*}{Highest Loss} & 30\% & 70.56 & 77.37  \\
 & 60\% & 67.17 & 69.07  \\ \hline
\end{tabular}
}
\end{table}


\input{bigtable_cifar.tex}

\subsubsection{Static \alg}

Here we present the results using the static policy termed \algns(5\%) and \algns(10\%) with constant $5\%$ and $10\%$ queries per batch arrival, respectively. 
We compare our \alg with the selection baselines from Section~\ref{ssec:baselines} with different number of active queries and noise rates.
Table~\ref{tab:bigtable_cifar} summarises the results.
As expected, \emph{No-Sel} which directly learns from the data has the worst results in the presence of noise, achieving at most $55.2\%$ accuracy.
However, our static \emph{BvSB} \alg with only $5\%$ active querying from the oracle comes remarkably close to \emph{Opt-Sel}, where the model is trained with only the clean samples from each batch.

Applying either sample selection or active querying alone achieves intermediate results. In particular with only active querying the results are close to \emph{No-Sel}. This is due to the small amount of samples actually relabeled by the oracle which is insufficient to combat the noise labels. Even if only noisy samples would be relabelled, the data would still contain a $25\%$ or $20\%$ of noisy labels for \emph{Al-only(5\%)} and \emph{Al-only(10\%)}, respectively. \emph{Q-only} alone is more efficient driving up the accuracy to approximately $65\%$ for CIFAR-10. However, \emph{Q-only} has very poor performance for CIFAR-100.
This comparison shows the effectiveness of both steps since neglecting any of both would result in a decrease in accuracy.

\subsubsection{Dynamic \alg}

We compare our dynamic policy based \algns$^{D}$ using a given budget of $B=2000$ with our static policy based \alg(5\%) that queries over the whole time horizon the same number of instances.
Figure~\ref{fig:cifar-10_static_dynamic_compare}(a) and (b) summarize the accuracy and loss results for CIFAR-10.
Using dynamic query allocation policy (green line) of the budget across the batches leads to a better performance $73.70\%$ than the static policy (orange line) $70.7\%$. Looking at the evolution, we observe a higher stability and less fluctuations based on noise. For reference, the red line shows the evolution under the perfect clean dataset. 



This result stems from the fact that the dynamic model queries more in the earlier batches when the model is less accurate and confident. Figure~\ref{fig:cifar-10_static_dynamic_compare}(c) shows the evolution of the number of active queries used in \algns$^{D}$ across the time periods. We se that indeed in the beginning the number of queries increases goes above the static assignment ($50$). In later batches the number of queries goes then near the same as the static case. Overall the budget is used over the whole time period. 

\subsubsection{Noise-resistant Model Baselines}
We compared our proposed \alg with the baselines described in Section~\ref{ssec:baselines}. Table~\ref{tab:baselines_table} summarizes the results for different noise-resistant model baselines and \alg.

For fair comparison, we apply use the same online stream data arrival pattern to train all models: each disposes of an initial clean dataset followed by and batches of stream data arrivals. Although these state-of-the-art models are successful in classification tasks of samples affected by label noise, they fail to adjust to the online setting. Here only a small portion of data is available in each time period for training. In the online scenario the best performance is achieved by \emph{Co-teaching} which however is still $15$ percent points lower than our \alg with $5\%$ active queries. Increasing the active queries to $10\%$ increases the gap by another percent point. The other four models, i.e. \emph{D2L}, \emph{Forward}, \emph{Bootstrap soft} and \emph{Bootstrap hard} all only achieve about $52\%$ accuracy. This underlines how our method copes very well with the online setting. Furthermore, some of mechanism proposed in these baselines could be coupled with our framework. For example, the loss correction implemented in \emph{Forward} could also be applied to our base classifier. We leave the investigation of such combinations for future work.



%% file: bigtable_cifar.tex
\begin{table*}[tp]
\caption{Summary of learning accuracy across different datasets and noise ratios. Five alternative approaches v.s. two different static versions of \alg, namely \algns(5\%) and \algns(10\%) using BvSB.}
\label{tab:bigtable_cifar}
\centering
\centering
\begin{tabular}{ccccccccc}

\hline
\multicolumn{1}{c|}{{Dataset}} &
\multicolumn{1}{c|}{\textbf{Noise}} & \multicolumn{1}{c|}{\textbf{Opt-sel}}    & \multicolumn{1}{c|}{\textbf{No-sel}} & \multicolumn{1}{c|}{\textbf{Q-only}}   & \multicolumn{1}{c|}{\textbf{AL-only($5\%$)}} & \multicolumn{1}{c|}{\textbf{AL-only($10\%$)}} & \multicolumn{1}{c|}{\textbf{\algns($5\%$)}} & \textbf{\algns($10\%$)} \\ \hline \hline

\multicolumn{1}{c|}{\multirow{2}{*}{CIFAR-10}} &
\multicolumn{1}{c|}{30\%}  & \multicolumn{1}{c|}{78.25} & \multicolumn{1}{c|}{54.92}  & \multicolumn{1}{c|}{65.32} & \multicolumn{1}{c|}{55.56}      & \multicolumn{1}{c|}{57.33}      & \multicolumn{1}{c|}{75.98} & \multicolumn{1}{c}{74.00} \\
\multicolumn{1}{c|}{} &
\multicolumn{1}{c|}{60\%}  & \multicolumn{1}{c|}{78.32}          & \multicolumn{1}{c|}{55.2}  & \multicolumn{1}{c|}{64.62} & \multicolumn{1}{c|}{56.04}      & \multicolumn{1}{c|}{56.11}      & \multicolumn{1}{c|}{70.71} & \multicolumn{1}{c}{74.07} \\ \hline

\multicolumn{1}{c|}{\multirow{2}{*}{CIFAR-100}} &
\multicolumn{1}{c|}{30\%}       & \multicolumn{1}{c|}{39.28}          & \multicolumn{1}{c|}{32.12}  & \multicolumn{1}{c|}{4.41} & \multicolumn{1}{c|}{32.10}      & \multicolumn{1}{c|}{32.14}      & \multicolumn{1}{c|}{39.10} & \multicolumn{1}{c}{36.55} \\
\multicolumn{1}{c|}{}  &
\multicolumn{1}{c|}{60\%}       & \multicolumn{1}{c|}{37.64} & \multicolumn{1}{c|}{14.48}  & \multicolumn{1}{c|}{4.15} & \multicolumn{1}{c|}{33.98}      & \multicolumn{1}{c|}{34.92}      & \multicolumn{1}{c|}{37.04} & \multicolumn{1}{c}{38.40} \\ \hline

\end{tabular}
\end{table*}